\documentclass[runningheads]{llncs}
\usepackage{graphicx}
\usepackage{amsmath}
\usepackage[utf8]{inputenc}
\usepackage[english]{babel}
\usepackage[dvipsnames]{xcolor}
\usepackage[misc,geometry]{ifsym}
\usepackage[backend=biber,
  style=lncs,
  sorting=none
]{biblatex}

\makeatletter
\newcommand{\printfnsymbol}[1]{%
  \textsuperscript{\@fnsymbol{#1}}%
}
\makeatother

\begin{document}
\title{Masked Image Modeling as a Framework for Self-Supervised Learning across Eye Movements}
\titlerunning{Self-Supervised Representation Learning via Saccadic Unmasking}

%
\author{Robin Weiler \inst{1,}\thanks{Equal contribution} \and Matthias Brucklacher \Letter\inst{1,}\printfnsymbol{1} \and Cyriel M. A. Pennartz \inst{1} \and Sander M. Boht\'e \inst{1, 2}}

\authorrunning{  }
%
\institute{Cognitive and Systems Neuroscience Group, University of Amsterdam, Science Park 904, 1098XH Amsterdam, Netherlands  \and
Machine Learning Group, CWI Amsterdam, Science Park 123, 1098 XG Amsterdam, Netherlands \\ 
\email{matthias.brucklacher@gmail.com}}

\maketitle              
\begin{abstract}
To make sense of their surroundings, intelligent systems must transform complex sensory inputs to structured codes that are reduced to task-relevant information such as object category. Biological agents achieve this in a largely autonomous manner, presumably via self-\allowbreak super-\allowbreak vised learning. Whereas previous attempts to model the underlying mechanisms were largely discriminative in nature, there is ample evidence that the brain employs a generative model of the world. Here, we propose that eye movements, in combination with the focused nature of primate vision, constitute a generative, self-supervised task of predicting and revealing visual information. We construct a proof-of-principle model starting from the framework of masked image modeling (MIM), a common approach in deep representation learning. To do so, we analyze how core components of MIM such as masking technique and data augmentation influence the formation of category-specific representations. This allows us not only to better understand the principles behind MIM, but to then reassemble a MIM more in line with the focused nature of biological perception. We find that MIM disentangles neurons in latent space without explicit regularization, a property that has been suggested to structure visual representations in primates. Together with previous findings of invariance learning, this highlights an interesting connection of MIM to latent regularization approaches for self-supervised learning. The source code is available under \url{https://github.com/RobinWeiler/FocusMIM} 

\end{abstract}

\section{Introduction}
\begin{figure}
\centering
\includegraphics[width=\textwidth]{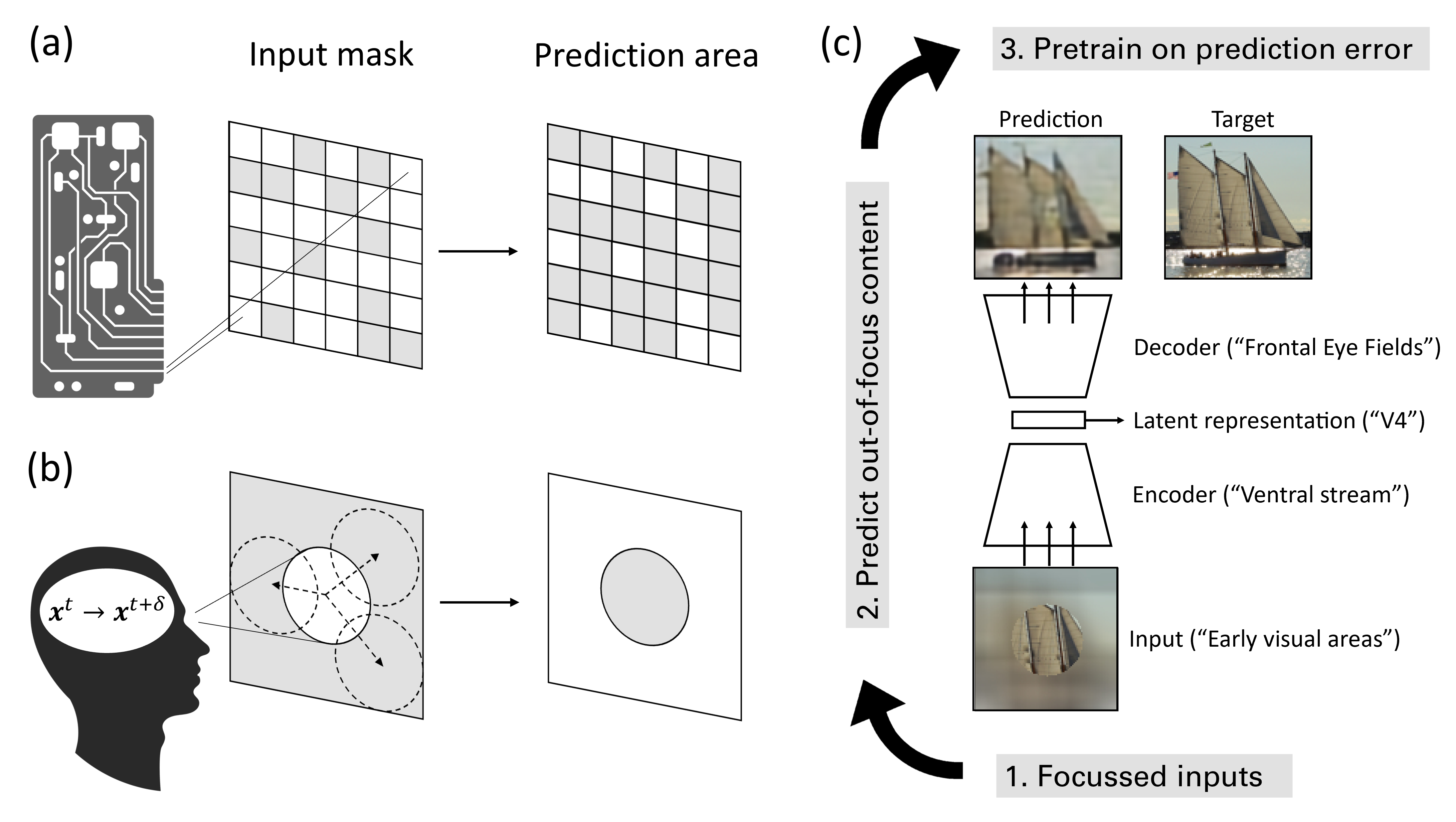}
\caption{Representation learning through eye movements. (a) Random patch masks used in artificial MIM approaches, gray patches are hidden from the network. (b) In our approach, selective masking is achieved through the inhomogeneous nature of foveal vision. Eye movements depicted as vectors reveal previously inaccessible or distorted information that is compared to the prediction $x^{t+\delta}$. (c) After pretraining on the prediction task, the quality of latent representations in the latent representation is assessed through classification accuracy in linear probing.} \label{fig:setting}
\end{figure}

Both biological and artificial intelligent systems must construct useful object representations (in the general case equivalent to classifiable) from large amounts of unlabeled data \cite{lecun2022path}. In self-supervised learning (SSL), two main approaches have crystalized over the last years: multi-view based learning (via contrasting \cite{oord2018representation}, latent regularization \cite{bardes2021vicreg} or distillation \cite{grill2020bootstrap}), and masked image-modeling (MIM) \cite{pathak2016context, he2022masked} which predicts occluded image content usually at the pixel-level, but to which also the more recently developed latent predictions \cite{assran2023self} can be counted. While constrastive methods have been suggested as a model of SSL in the brain, structuring representational geometry via temporal \cite{halvagal2023combination} and spatial cues \cite{illing2021local}, we postulate that the brain may be engaged in MIM via eye movements and attention shifts (Fig.~\ref{fig:setting}): Perception, at a singular point in time, is selective, both through foveal vision and selective attention \cite{simons1999gorillas}. Sequentially, saccadic eye movements and attention shifts then flood the sensory stream with new information - in a manner that is to some degree predictable from knowledge about the direction and magnitude of the gaze shift (via a corollary discharge \cite{crapse2008frontal, von1950reafferenzprinzip}) and image structure. \\

Studying the principles behind MIM is attractive, as it requires fewer assumptions about temporal sequences and data augmentations than the mentioned multi-view approaches to biological SSL and achieves performance on par or close to contrastive methods \cite{woo2023convnext}. MIM fits well with predictive theories of cortical processing \cite{dayan1995helmholtz, rao1999predictive, friston2005theory}. Both conceptualize perception as the process of iden-\\

\noindent
 tifying a hierarchical latent variable model that could have caused the sensory inputs \cite{pennartz2015brain, kong2023understanding} (this generative aspect is lacking in discriminative SSL models) and are predictive of future inputs. Experimental evidence for predictions of visual content across saccades comes, e.g., from \cite{ehinger2015predictions}. Anatomically, the frontal eye fields have been proposed to predict across eye movements to stabilize perception \cite{crapse2008frontal}. We postulate a larger role of these predictions: the learning of visual features useful for downstream tasks such as classification. \\

Here, we investigate the influence of multiple variations to MIM in convolutional neural networks (CNNs) that align well with biological constraints. First, the influence of different masking strategies is tested, resulting in a peripheral masking strategy that learns strong representations and is more in line with the focused nature of biological perception. In doing so, we discover that representations are markedly improved, and that neurons become more decorrelated, when peripheral information is completely suppressed during learning. Second, the influence of data augmentation is tested, showing that these are especially relevant for the peripheral masking strategy. Third, we show that the relevant loss signal originates from the main object: although networks reconstruct the context too, this is not necessary for representation learning. In fact, we find that under some conditions networks learn faster when being only rewarded for predicting the foreground, which is interesting as segmentation may naturally be available from geometric cues or motion \cite{tsao2022topological, brucklacher2023learning}. Indeed, availability of segmentation has been shown to aid constrastive SSL if leveraged during pretraining \cite{henaff2021efficient}, while its effect on MIM has previously not been studied to our knowledge. \\

\section{Related Work}
The breakthrough for MIM in representation learning arguably was achieved using random patch masks (Fig.~\ref{fig:setting}a) in a vision transformer \cite{he2022masked}, an architecture that differs from information processing in biological networks. Investigations into the influence of masking strategies found that the optimal masking ratio falls roughly between 40\% and 80\%, being sufficiently large but not overly aggressive \cite{kong2023understanding, he2022masked}. A more biologically compatible architecture (CNNs) for MIM was used by Pathak et al. \cite{pathak2016context}, as do we for this reason. Pathak et al. found that a central mask (predicting from peripheral context to a small central region) performed worse than randomly distributed patch or region masks, although quantitative reports were not provided. Here, we address this shortcoming and systematically study the choice of the masking paradigm. A recent CNN-based MIM is ConvNeXt-v2 \cite{woo2023convnext}, which utilizes random patch masks like the transformer-based methods in combination with a shallow decoder, whereas we strive to make the masking paradigm more plausible. Xie et al. \cite{xie2022simmim} further demonstrated the potential for MIM in CNNs, achieving high accuracy in a ResNet-based model. In contrast to their evaluation paradigm that employs the same (full) dataset both during pretraining and fine-tuning \cite[see also ][]{woo2023convnext}, we return to the ethologically relevant setting with limited amounts of labeled data. \\

Biological approaches to SSL derive mostly from multi-view approaches. They avoid representation collapse, a catastrophic network state in which different inputs are projected onto the same latent activity pattern and thus cannot be distinguished anymore, by introducing an expanding force that pushes representations apart. Illing et al. \cite{illing2021local} used a single negative sample at a time, interpreted as being obtained by larger eye movements. Their approach converts contrastive predictive coding \cite{oord2018representation} into a fully local learning rule. Interestingly, the model also has a generative component (predicting the next representation in a sequence of views) as a means to representation learning that is not investigated further. Local predictive learning (LPL) \cite{halvagal2023combination}, on the other hand, relies on pushing representations apart based on their temporal distance, converting variance regularization \cite{bardes2021vicreg} into a local learning rule. In contrast, MIM, as we employ it, prevents representation collapse in a more minimalist way, i.e., without the necessity of negative examples \cite{illing2021local} or accumulated temporal statistics \cite{halvagal2023combination}, but instead solves a generative problem in which loss minimization is incompatible with a collapsed solution. As the approach presented here is an initial step, we use the spatially non-local backpropagation, however, see \cite{lillicrap2020backpropagation, whittington2017approximation} and subsequent work for biologically plausible solutions yielding comparable weight updates. \\

Previous functional models have combined focused vision with eye movements in a variety of ways, from controlling information gain \cite{renninger2004information} and active inference \cite{friston2012perceptions}, to achieving representation invariance and a reduced need for detail in generative model learning  \parencite{larochelle2010learning}. The model of Thompson et al. \cite{thompson2022learning} combines spatially structured eye movements with visual information processing in an enactive approach to counting. The approach of Illing et al. \cite{illing2021local} uses on- vs. off-object eye movements as a heuristic to shape representational geometry in latent space. In comparison, our approach is more reduced in dynamics and focuses instead on the consequences for neural representation learning when casting eye movements as a self-supervised predictive task. It should also be noted that all mentioned functional interpretations are mutually compatible.

\section{Method}

\subsection{Model}
We reduce the task of predicting across eye movements to its essence and avoid learning a remapping of spatial receptive fields, which we argue is orthogonal to the process of information restoration that drives representation learning \cite{kong2023understanding}. Thus, inputs and predictions take place in the same reference frame (Fig.~\ref{fig:setting}c), as in other MIM approaches \cite{he2022masked, woo2023convnext, kong2023understanding}. The network architecture, shown in Fig.~\ref{fig:setting}c, resembles an autoencoder. Unless explicitly noted otherwise, we use the following setup: The encoder reduces the spatial dimension from 96\,x\,96 (with three color channels) to 12\,x\,12\,(x\,128 channels). Due to the large receptive fields of the latent space neurons at the output of the encoder, they map roughly onto neurons in visual area V4 or higher in primates \cite{smith2001estimating}. The inversely structured decoder (with deconvolutions instead of convolutions) maps the latent representations back onto a 96\,x\,96\,x\,3 output, constituting the model's expectation of image content. Both encoder and decoder comprise nine ResNet18-style blocks \cite{he2016deep} of two convolutions passed by a skip connection. Convolutions are followed by a ReLU activation function except the last decoder layer, which is followed by a sigmoid function. The convolution at the entry to each third block has stride\,=\,2, which is where we double and half the number of channels in the encoder and decoder, respectively, starting with 32 channels at the first encoder block.

\subsection{Self-Supervised Pretraining}
\label{sec:methods-pretraining}
\begin{figure}
\centering
\includegraphics[width=0.8\textwidth]{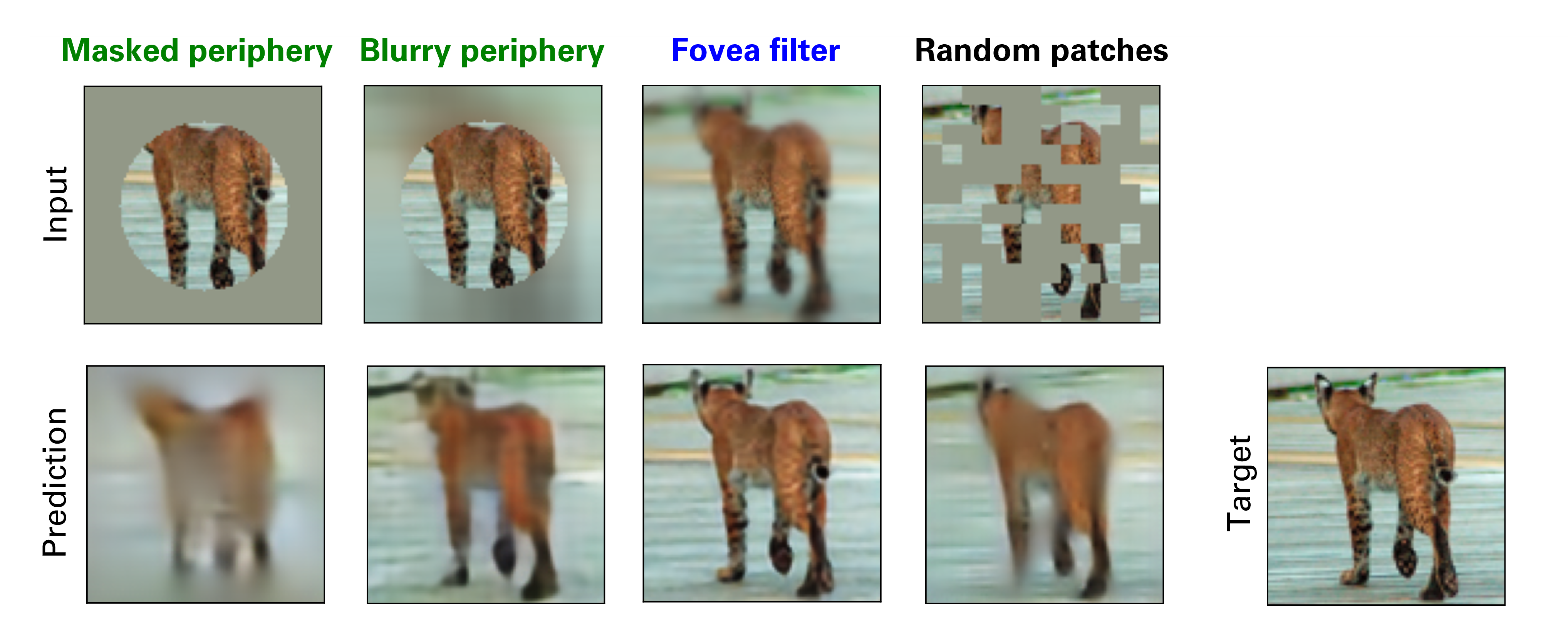}
\caption{The masking strategies define the pretraining task. Top row: Exemplary inputs for different pretraining paradigms, with masked areas covered in the image's gray average color. Masked periphery and random patches are shown with 60\% masking ratio. Bottom row: Predictions from the given inputs after pretraining. Right image: Ground truth} \label{fig:reconstructions}
\end{figure}

As other work in MIM, we define pretraining as the self-supervised training phase in which the model is presented with a partial view of an image and attempts to predict the image pixel by pixel in full resolution. The loss function is given as the mean squared reconstruction error in parts of visual space with incomplete inputs (masked or blurred, depending on the paradigm), normalized by the area over which the loss is calculated. This generative loss is reminiscent of predictive coding \cite{rao1999predictive}. Pretraining was conducted on the unlabeled split of the STL-10 dataset \cite{coates2011analysis}, which was specifically constructed for SSL with few labeled examples, and is sufficiently naturalistic and large to exclude the need for another dataset. We investigated a broad spectrum of input masking strategies (Fig.~\ref{fig:reconstructions}) that pertain to the original hypothesis of peripheral masking: \\

\noindent\textbf{Random patches}: The most common masking strategy, used, e.g., in \cite{he2016deep, woo2023convnext, kong2023understanding} (cf. Fig.~\ref{fig:setting}a) serves as an optimal baseline. We split the image into 144 patches of 8\,x\,8 pixels and replaced the masked image content with the average color of the image.\\
\newpage

\noindent \textbf{Masked periphery}: A central circle in full resolution, with the periphery masked with the sample's average color (see also Fig.~\ref{fig:setting}b). Biologically, this strict discarding of out-of-focus information is imaginable through inattentiveness \cite{simons1999gorillas}.\\

\noindent\textbf{Blurry periphery} and \textbf{Blurry random patches}: Variations of the above paradigms with local Gaussian blurring instead of a uniformly colored mask to test the influence of incomplete masking.\\

\noindent\textbf{Foveal filter}: A centered circle with a small radius in full resolution. Outside the circle, the level of blur increases with distance to the center, mimicking the physiological constraints of human foveal vision, i.e., the distribution of cone photoreceptors across the retina \cite[p.~244]{purves2004neuroscience}. The implementation follows \cite{perry2002gaze, jiang2015salicon}. \\

For both random patches and masked periphery (using a square mask instead of a circle), we experimented with sparse convolutions as in \cite{woo2023convnext} to avoid processing of borders between masks and image content. However, we found no significant difference in linear probing accuracy (as described below), so we continued with standard convolutions.

\subsection{Choice of Pretraining Hyperparameters}
During pretraining, we employed random-resized cropping (scale\,$\in$\,$\{0.08, 1.0\}$, aspect ratio\,$\in$\,$ \{0.75, 1.33\}$). We used the Adam optimizer \cite{kingma2014adam} with weight decay\,=\,1e$-8$, batch size\,=\,512, and selected the learning rate from \{5e-5, 1e-4, 5e-4\} that led to the best classification performance (described below) for each of the major pretraining tasks (masked periphery, random patches, foveal filter) to render the obtained findings independent of the choice. This value was then kept when blurry information was introduced. For each of the major masking paradigms, we additionally varied the masking ratio from 10\,-\,90\% in steps of 10 to choose a model with optimal classification accuracy. For the random patches, we found a masking ratio of 60\% to lead to the best representations. For the masked periphery, a higher ratio of 80\% performed slightly better. We thus continued with this fraction for the remaining results, and refer to \cite{kong2023understanding, he2016deep} for more detailed studies on the influence of the masking ratio. After 500 epochs of pretraining, we continued with the model that achieved the lowest reconstruction loss for further analysis.

\subsection{Linear Probing to Assess Representation Quality}
To analyze the quality of the learned representations after pretraining, linear probing was conducted. Here, we followed the standard SSL procedure of discarding the decoder and appending a linear layer to the frozen encoder. The 128\,x\,12\,x\,12\,x\,10 (latent dimension\,\textasteriskcentered{}\,classes of the STL-10 labeled splits) parameters of the linear layer were then fit without further augmentations on the training split of the STL-10 dataset (5,000 images), of which we used 1,000 for training validation and early stopping at the minimum validation loss. Irrespective of pretraining paradigm, we used images without mask or blurring in this phase, as these consistently led to higher readout accuracy. As in pretraining, we employed the Adam optimizer \cite{kingma2014adam} (learning rate\,=\,1e-4, weight decay\,=\,1e-8, batch size\,=\,512). Classification accuracy was then evaluated on the 8,000 test images, averaged across five randomly seeded runs. We also experimented with multilayer perceptron readout and fine-tuning (retraining the whole network), but neither of these approaches led to large differences in accuracy.

\subsection{Generalization to a different network architecture}
To test in how far the influence of masking strategy depends on the network architecture, we pretrained a modern CNN, the ConvNeXt-v2-Pico \cite{woo2023convnext}. Key differences to the ResNet described above are increased receptive field size, separate depthwise and pointwise convolutions, the use of GELU activation functions, and a shallow decoder consisting of a single block. Pretraining on the STL-10 dataset was conducted in two configurations: with random patch masks or with masked periphery. Due to the ConvNeXt-v2's initial stage that splits the input image into 7\,x\,7 patches, we implemented peripheral masking as predicting from the central 3\,x\,3 patches to the surround. For the random patches, we used a masking ratio of 60\%, as suggested in \cite{woo2023convnext}. Models were then pretrained for 370 epochs with hyperparameters close to \cite{woo2023convnext}: batch size\,=\,512, learning rate selected for optimal performance out of \{1.6e-4, 5e-4, 1.5e-3\}, AdamW optimizer \cite{loshchilov2017decoupled} with 20 epoch linear warmup and subsequent cosine decay, weight decay\,=\,0.05 as well as random resized cropping and horizontal flipping augmentations. For linear probing and fine-tuning, random erasing was additionally applied, and warmup was reduced to three epochs, with a base learning rate of 0.0006 (fine-tuning) and 0.008 (linear probing). While fine-tuning led to the best performance when using global average pooling before the classification head, linear readout profited from skipping this, so we proceeded with the optimal settings for each. 

\subsection{Correlation of Neurons in Latent Space}
\label{sec:methods-covariance}
In addition to linear discriminability, representations can also be analyzed based on their statistical properties. One desirable property here is decorrelation \cite{barlow1961possible, bardes2021vicreg, zbontar2021barlow}. To quantify this, the covariance matrix is required \cite{bardes2021vicreg, zbontar2021barlow}: 
\begin{equation}
    C(R)=\frac{1}{n-1} \sum_{i=1}^n\mathbf{r}_i^{ }\mathbf{r}_i^T 
\end{equation}
with $R$ defined as a tensor containing the $n$ latent representation vectors $\mathbf{r}_i$ that are z-scored across the batch dimension, with $n$ equal to the number of samples (we compute across one mini-batch with $n=512$). As in \cite{bardes2021vicreg}, the sum of the squared off-diagonal entries in $C$ then quantifies the degree to which neurons are correlated:
\begin{equation}
\label{eq:covariance}
    c(R) = \sum_{i\neq j }[C(R)]^2_{i, j}
\end{equation}
\begin{figure}
\centering
\includegraphics[width=0.77\textwidth]{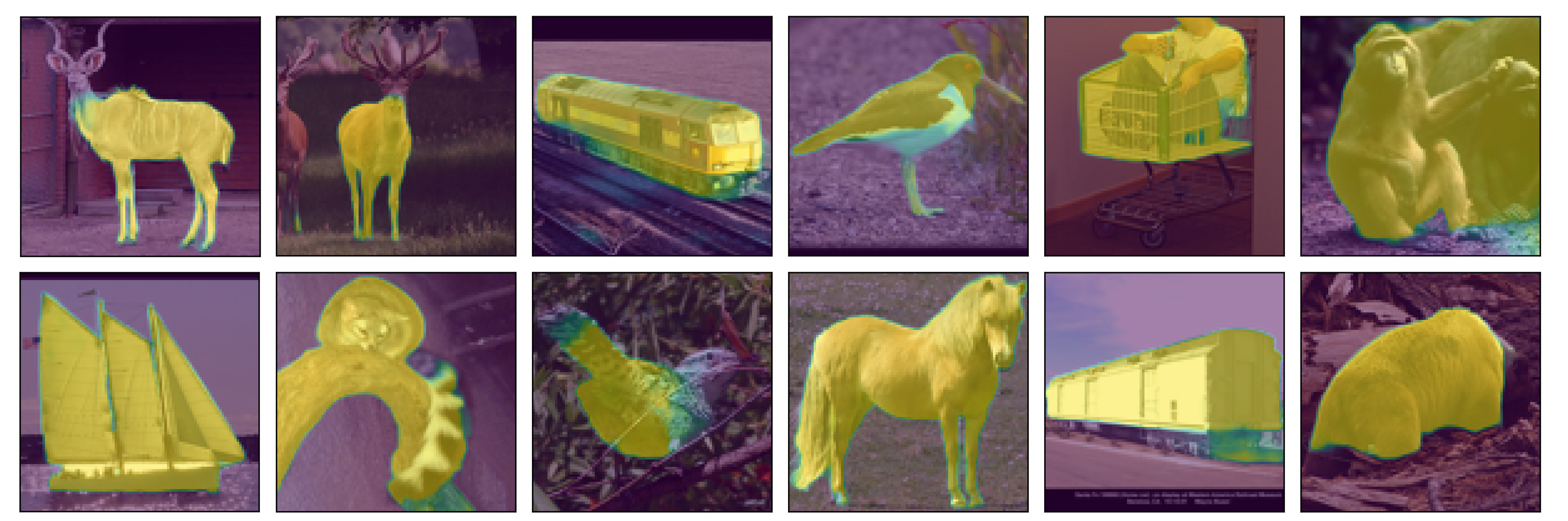}
\caption{Representative presegmentation masks used to investigate how discarding loss signals from the background affects network performance. The masks were obtained with the \textit{rembg}-library, with brightness indicating confidence, overlaid onto the respective STL-10 images. }\label{fig:presegmentation_masks}
\end{figure}
\subsection{Disentangling Predictions of Object vs. Background}
\label{sec:methods-presegmentation}
Unlike assumed in section \ref{sec:methods-covariance}, eye movements reveal information sequentially, often targeting salient points such as objects \cite{hayhoe2005eye, yarbus1967eye}. Objects may also be segmented already before recognition, either from motion signals \cite{brucklacher2023learning, tsao2022topological} or occlusion geometry \cite{tsao2022topological}, making them prioritized prediction targets. We thus evaluated the influence of reduced learning signal availability from the background. First, we obtained a figure-background segmentation mask using the \textit{rembg}-library (\url{https://github.com/danielgatis/rembg}). We discarded $\sim$\,8,000 images for which no clear segmentation mask could be obtained. In these, less than 100 pixels (out of 96\,x\,96) received a confidence score of 0.8 (on a scale from 0 to 1) or higher. In the remaining images, the masks were generally high-quality and only small object parts were missed (Fig.~\ref{fig:presegmentation_masks}). During pretraining, the reconstruction loss was then multiplied with the foreground confidence score at each point in space and normalized to compensate the reduced prediction area. \\

\section{Results}
To identify the minimally necessary components for biological MIM, we applied various combinations of masking strategy and data augmentation, and plot their influence on representation quality in Fig.~\ref{fig:accuracy_covariance}a-c.

\subsection{Influence of Masking Strategy}
Out of the used masking strategies, we identified the masked periphery condition as the most promising candidate for a biological model. This masking technique inspired by foveal visual perception led to the second-highest classification accuracy in linear readout (mean value $67.9\pm0.4\%$) when crop-and-resize transformations were used during pretraining, outperformed only by the random patch masks ($70.2\pm0.4\%$). Similarly, in the ConvNeXt-v2 architecture, peripheral masking led to a high linear readout accuracy of $71.6\pm0.6\%$ (fine-tuning:

\newpage
\begin{figure}
\centering
\includegraphics[width=1.\textwidth]{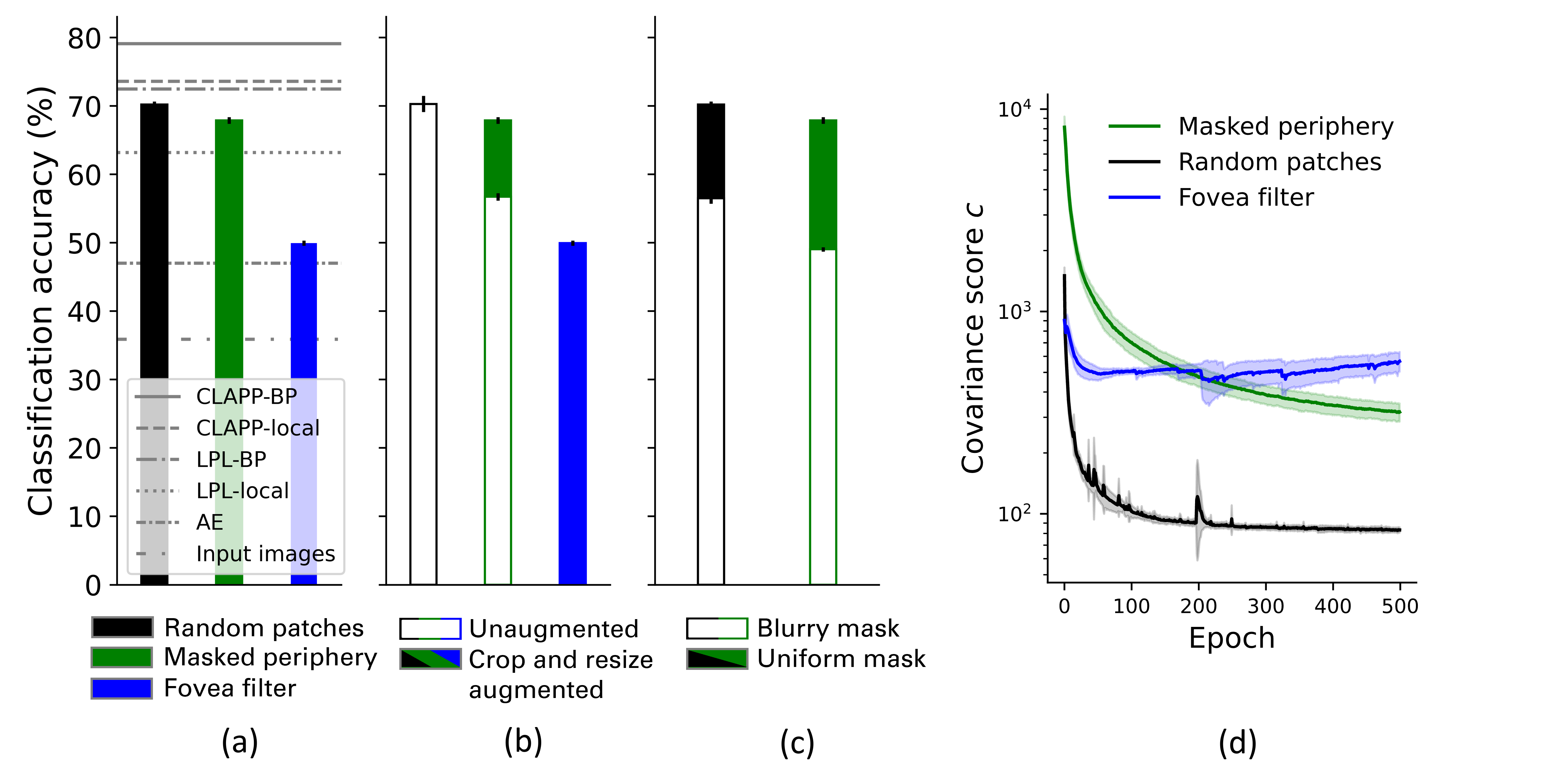}
\caption{Linear readout accuracy quantifies representation quality after pretraining. a) Influence of masking strategy in comparison to the multi-view approaches CLAPP \cite{illing2021local} and LPL \cite{halvagal2023combination}, that both come in two variants trained with a local learning rule or backpropagation. The remaining baselines are pure autoencoding (AE) from full image to full image in the same network as the masked methods, and directly conducting linear probing on the input images. b) Influence of data augmentation. In the left- and rightmost bar, the filled and outlined bars overlap, i.e., classification accuracy is unaffected by augmentation. c) Incomplete masking by Gaussian blurring, instead of uniform average coloring, drastically decreased representation quality. d) Masked image modeling implicitly decorrelates latent space neurons. Error bars and shaded regions indicate the standard deviation across five randomly seeded runs.} 
\label{fig:accuracy_covariance}
\end{figure}
\noindent
$90.9\pm0.3\%$), compared to $68.2\pm0.3\%$ (fine-tuning: $89.4\pm0.4\%$) attained with random patch masks. Strong performance in these settings supports the assumption that solving the pretraining tasks of predicting hidden image areas leads to discovery of object representations \cite{kong2023understanding}. In contrast, deblurring the peripheral region from the foveal filter resulted in comparably poor representations, with readout accuracy not much higher than pure autoencoding. Presumably, this condition allows the network to find a local solution that does not require global integration \cite{kong2023understanding}. To more closely investigate the effect of input information content on classification accuracy, we replaced uniform masking with Gaussian blurring (Fig.~\ref{fig:reconstructions}, third column). Despite the low information content, the introduced information was sufficient to substantially degrade performance in both the masked periphery and random patch condition (Fig.~\ref{fig:accuracy_covariance}c).

\subsection{Decorrelation of Latent Space Neurons}
To investigate the degree to which neurons in latent space develop independent tuning properties, we computed the correlation term $c$ using Eq. \ref{eq:covariance}. Interestingly, both masked periphery and random patches decorrelated neural representations (Fig.~\ref{fig:accuracy_covariance}d), and more strongly than the foveal filter. This decorrelation, reminiscent of sparse coding \cite{olshausen1996emergence}, emerged without being explicitly enforced as in other approaches \cite{olshausen1996emergence, bardes2021vicreg, zbontar2021barlow}.

\subsection{Influence of Data Augmentation}
Applying crop-and-resize augmentations during pretraining proved necessary for representation learning in the masked periphery condition. When leaving out this augmentation, readout accuracy dropped from $67.9\pm0.4\%$ to $56.7\pm0.5\%$ (Fig.~\ref{fig:accuracy_covariance}b). Biologically, cropping and resizing can be approximately obtained by changes in viewpoint and viewing angle, especially since no assumption about temporal co-occurrence is made: differently augmented versions of an image are presented in different epochs. Interestingly, neither the foveal filter nor the random patches were affected by ablating the crop-and-resize transformation. Maintaining the position of the patch masks across epochs alone did not reduce readout accuracy ($70.3\pm0.9\%$), but in combination with ablation of the crop-and-resize augmentation, accuracy reduced to $62.5\pm0.8\%$. We conclude that the model requires exposure to a variety of prediction tasks from each object, whether through data augmentation or through changes in mask position.

\subsection{Predicting Object vs. Background}
\begin{figure}
\centering
\includegraphics[width=0.6\textwidth]{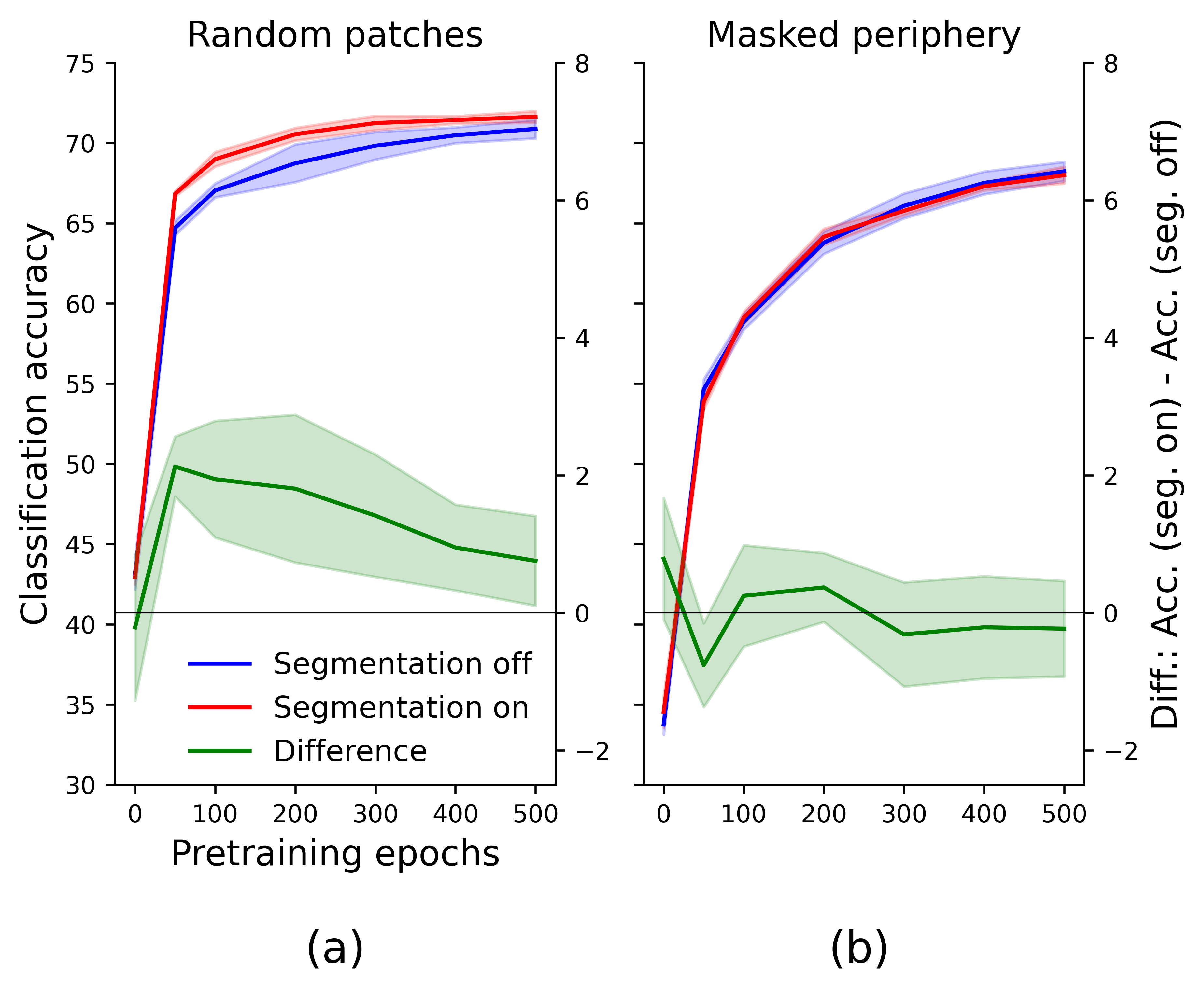}
\caption{Influence of presegmentation and masking strategy on pretraining efficiency. (a) When using random patches, using presegmentation to weight the reconstruction loss accelerated pretraining. (b) This was not the case when restricting the reconstruction loss to the foreground object. Shaded regions indicate the standard deviation across five randomly seeded runs.} 
\label{fig:presegmentation_plot}
\end{figure}

Restricting the pretraining loss to the main object instead of the background (cf. section \ref{sec:methods-presegmentation}) did not affect linear probing accuracy negatively, suggesting that on-object predictions are sufficient for classification. While there was no substantial difference in readout accuracy when disregarding, as opposed to when using, loss signal from the background in the masked periphery condition (Fig.~\ref{fig:presegmentation_plot}b), the network pretrained using random patch masks profited from disregarding the background, especially in early epochs (Fig.~\ref{fig:presegmentation_plot}a). 

\subsection{Reconstruction Quality}
Beyond the network's discriminative function, its generative capacity bears functional relevance, e.g., for counterfactual reasoning, predictions of occluded image content, and imagination \cite{pennartz2015brain}. An interesting effect from generative model learning could be observed in the networks pretrained with uniform (average colored) and blurry masks: These networks also reconstructed in locations in which no loss was calculated. While this would be expected to some degree when the masked area varies across image presentations (random patches), it is surprising in the masked periphery condition (Fig.~\ref{fig:reconstructions}, second column). Here, also the central region was reconstructed reasonably well, although it did not contribute to the pretraining loss. This holistic percept fits well with proposals on the role of visual predictions in creating a stable visual representation \cite{crapse2008frontal, pennartz2015brain}. \\

Across masking paradigms, reconstruction accuracy was uncorrelated with readout accuracy: while the foveal filter achieved the best reconstructions (Fig.~\ref{fig:reconstructions}), it performed poorly in linear probing (Fig.~\ref{fig:accuracy_covariance}a). Given a pretraining paradigm, however, longer training on the reconstruction task also improved readout (Fig.~\ref{fig:presegmentation_plot}). \\

\begin{figure}
\centering
\includegraphics[width=1.0\textwidth]{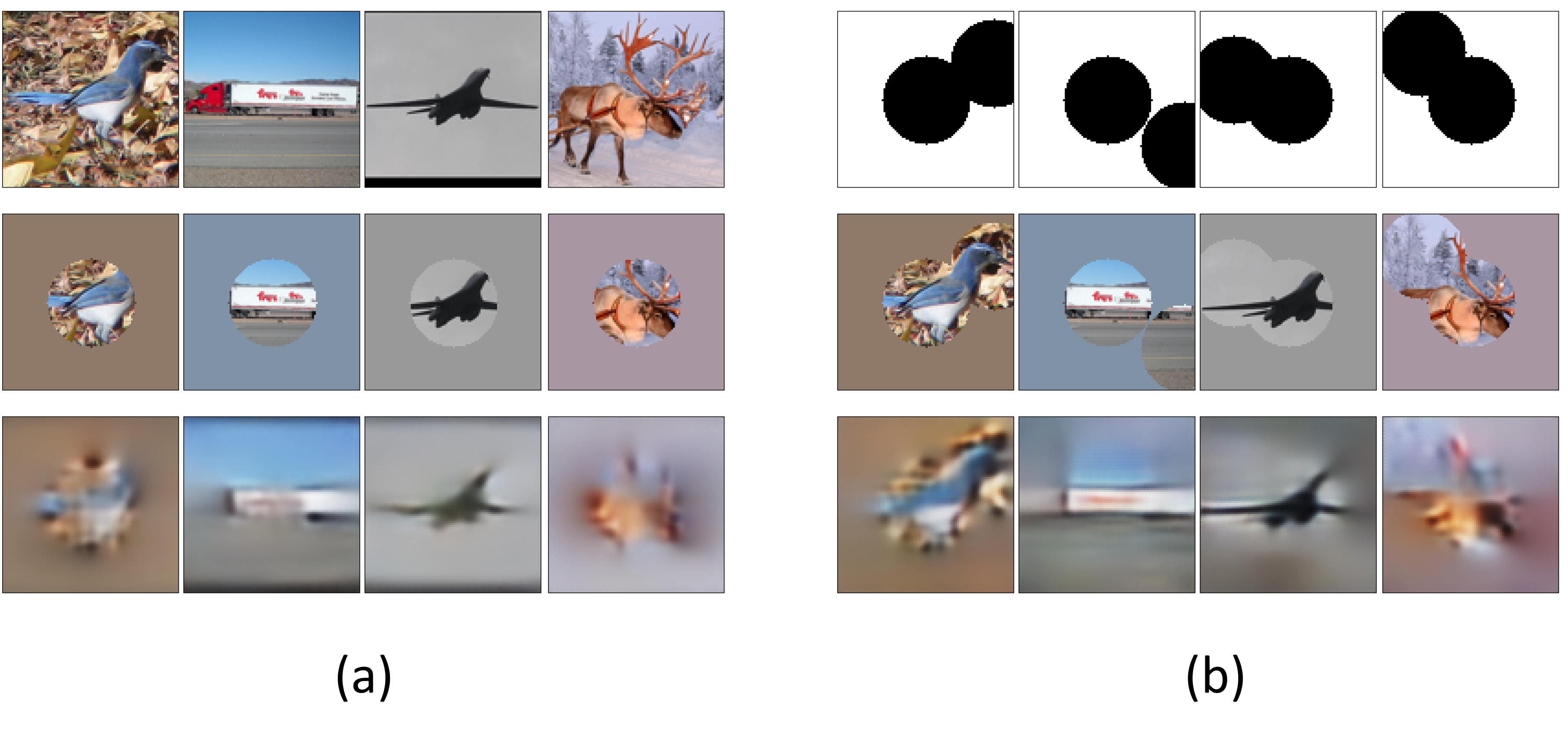}
\caption{Sampling via covert attention improves predictions and stabilizes perception. (a) Top row: ground truth. Middle row: input in the masked periphery condition with a masking ratio of 80\%. Bottom row: reconstructions after pretraining. (b) Same as (a) with additionally provided information via a second, non-central input circle (covert attention). The top row depicts the input mask consisting of these two circular areas (fovea and attention).} \label{fig:attention}
\end{figure}

A third angle to analyze the generative capacity of the network stems from the fact that eye movements are often preceded by covert attention shifts, which increase information sampling from these areas \cite{houtkamp2003gradual}. The sampled information can then be used to improve predictions about the expected image content. To test whether the model could exploit such a mechanism for improved reconstructions, we provided the model with a second circular input sampling area (covert attention) on top of the peripheral masking while constraining the error to the foreground (section \ref{sec:methods-presegmentation}). The additionally provided information indeed led to more truthful predictions (Fig.~\ref{fig:attention}), i.e., reduced prediction errors (from 2.6e-2 to 2.3e-2), which ties the approach back to earlier theories of predictions as a stabilizing mechanism \cite{crapse2008frontal}. 
Applying this twofold sampling during pretraining, however, did not alter readout accuracy significantly, irrespective of whether the covert attention circle was placed at random or on the main object.

\section{Conclusion}
In this paper, we investigated key components of masked image modeling (MIM) through the lens of biological vision. As a similarly performant alternative to the random patch masks common in artificial representation learning \cite{he2022masked, woo2023convnext}, we identified the more biologically plausible peripheral masking (Fig.~\ref{fig:accuracy_covariance}a). Provided with differently augmented views of the input images (Fig.~\ref{fig:accuracy_covariance}b), this approach does not pose restrictions on the temporal sequence in which inputs are encountered, as does \cite{halvagal2023combination}, or on knowledge about the spatial extent of an object to obtain negative samples, as does \cite{illing2021local}. We conclude that peripheral masking and prediction is a candidate mechanism for self-supervised learning in the brain, with saccadic prediction errors computed in the early (but likely not primary) visual cortex \cite{ehinger2015predictions} where bottom-up signals are compared to feedback from the frontal eye fields \cite{crapse2008frontal}. However, further experimental studies are required to definitely link these error signals to plasticity in the ventral stream, and to elucidate whether covert attention shifts can elicit similar prediction errors. Functionally, an additional advantage of the employed approach in contrast to purely discriminative methods is the acquisition of a generative model (Fig.~\ref{fig:reconstructions}, Fig.~\ref{fig:attention}). That multi-view SSL has been implemented with more local learning rules than backpropagation \cite{illing2021local, halvagal2023combination,lowe2019putting, siddiqui2023blockwise} prompts the question of whether the same can be done with MIM. From a theoretical perspective, implicit decorrelation of latent space neurons (Fig.~\ref{fig:accuracy_covariance}d), together with previous observations of implicit invariance learning \cite{kong2023understanding2}, connect the MIM framework to SSL via latent regularization \cite{bardes2021vicreg}, where these objectives are explicitly enforced. This raises the question of whether and how synergies could be efficiently exploited in models combining both learning paradigms \cite{assran2023self, jing2022masked}. As a general requirement for high classification accuracy across all investigated masking strategies, we identified opacity of the input masks (Fig.~\ref{fig:accuracy_covariance}c), just as humans sometimes do not perceive objects in their field of view when attending to another point in space \cite{simons1999gorillas}. As primates employ eye movements in a strategic and enactive way \cite{thompson2022learning}, an interesting extension would be to learn from multiple peripherally masked glimpses in a sequence through spatiotemporally masked videos \cite{feichtenhofer2022masked}. 

\subsubsection*{Acknowledgements}
This project has received funding from the European Union’s Horizon 2020 Framework Programme for Research and Innovation under the Specific Grant Agreement No. 945539 (Human Brain Project SGA3 to C.P. and S.B.).
We thank SURF (\url{www.surf.nl}) for the support in using the National Supercomputer Snellius, especially Thomas von Osch for the discussions. 

\printbibliography

\end{document}